\documentclass[letterpaper]{article}

\usepackage{natbib,alifeconf}  

\usepackage{placeins}
\usepackage{algorithm}
\usepackage[noend]{algpseudocode}
\usepackage{mathtools}

\title{Adapting to Unseen Environments through\\Explicit Representation of Context}
\author{Cem Tutum$^{1}$ \and Risto Miikkulainen$^{1, 2}$ \\
\mbox{}\\
$^{1}$The University of Texas at Austin, Austin, TX 78712 \\
$^{2}$Cognizant Technology Solutions, San Francisco, CA 94111 \\
tutum@cs.utexas.edu} 

\begin{document}
\maketitle

\begin{abstract}
In order to deploy autonomous agents to domains such as autonomous driving, infrastructure management, health care, and finance, they must be able to adapt safely to unseen situations. The current approach in constructing such agents is to try to include as much variation into training as possible, and then generalize within the possible variations. This paper proposes a principled approach where a context module is coevolved with a skill module. The context module recognizes the variation and modulates the skill module so that the entire system performs well in unseen situations. The approach is evaluated in a challenging version of the Flappy Bird game where the effects of the actions vary over time.  The Context+Skill approach leads to significantly more robust behavior in environments with previously unseen effects. Such a principled generalization ability is essential in deploying autonomous agents in real world tasks, and can serve as a foundation for continual learning as well.
\end{abstract}

\section{Introduction}

Generalization to unseen situations is an important capability for autonomous agents. Especially in real-world decision making and control applications such as autonomous driving, robotics, process control, health care, and finance, the agents routinely need to adapt safely to unseen situations. A common practice is to train these models, mostly deep neural networks, with the data collected from a limited number of hand-designed scenarios. However, the tasks are often too complex to anticipate every possible scenario, and this approach is not scalable. Moreover, these models can be brittle when they are exposed to even small variations or noise.

One popular approach to address this problem is few-shot learning, in particular metalearning, either by utilizing gradients \citep{Schmidhuber87_PhD, Thrun98_Learn, Finn17_MAML} or evolutionary procedures \citep{Fernando18_Baldwin, Grbic19_EvoMetaLearn}. In metalearning, systems are trained by exposing them to a large number of tasks, and then tested for their ability to learn new relevant but unseen tasks. There are also a number of approaches mostly for supervised learning setting where new labels need to be predicted based on limited number of training data. However, applications in control and decision making, including reinforcement learning problems, are very limited \citep{Kansky17_Schema}. 

The approach in this paper is motivated by prior work on opponent modeling in poker \citep{Li17_Poker,li:gecco18}. In that domain, an effective approach was to evolve one neural network, the game module, to decide what move to make, and another, the opponent module, to monitor the opponent, and modulate those decisions by taking the opponents playing style into account. When trained with only a small number of very simple but different opponents, the approach was able to generalize and play well against a wide array of opponents, include some that were much better than anything seen during training.

In a sense, the opponent forms a context for the decision making in poker. Each decision needs to take into account how the opponent is likely to respond, and select the right action accordingly. The player can thus adapt to many different game playing situations immediately, even those that have not been encountered before. In this paper, this approach is generalized and applied to control and decision making more broadly. In more general terms, a skill network reacts to the current situation, and a context network integrates observations over a longer time period. A third, controller, network combines the outputs of both networks, thus modulating decision making through context. Such a Context+Skill system can thus generalize to more situations than any of its components alone.

The Context+Skill approach is evaluated in this paper on an extended version of the popular Flappy Bird game. This version includes more actions and physical effects (i.e.\ forward flap and drag in addition to flap up and gravity). Such an extension allows generating a range of unseen scenarios both by extending the range of effects of those actions as well as their combinations. The approach generalizes remarkably well to new situations, and does so much better than its components alone. Context+Skill approach is thus a promising approach for building robust autonomous agents in real-world domains.

The remaining sections of this paper are organized as follows: Methodology section presents the experimental set up and the test domain, the architecture of the neural networks, and the multiobjective evolution procedure for constructing the system. Next, learning and generalization results are presented, demonstrating that the Context+Skill approach indeed performs better than its parts. The behaviors of these networks are contrasted in the Behavior Analysis section, finding that Context+Skill can anticipate results of its actions more accurately, making it possible to adapt to unseen situations. 


\section{Methodology} \label{sec_method}

This section introduces the experimental setup, the neural networks used as the control policies for the agent, and the evolutionary training methodology.

\subsection{The Flappy Ball Domain} \label{Game}

\begin{figure}
\centering
\includegraphics[width=2in]{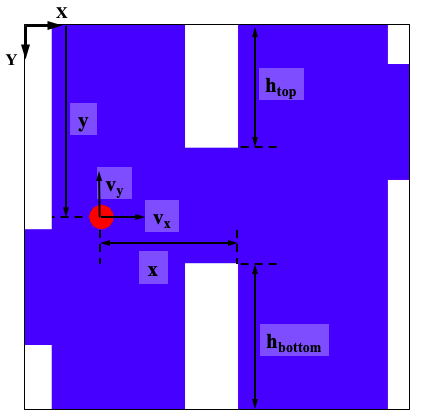}
\caption{A scene from the Flappy Ball game. The red circle represents the agent and the white columns are pipes that move from right to left as the game progresses. At each time step, the agent can flap up or forward or both; if it does not, gravity will pull it down and drag will slow it down. The origin of the coordinate system (0,0) is at the upper left corner; thus the action of flapping up results in values of negative y-velocity or smaller y-position. The six variables identified in the figure constitute the input information that the agent receives at each time step.}
\label{fig_FlappyBall}
\end{figure}

Flappy Ball is an extension of the popular Flappy Bird computer game \citep{FlappyBird_wiki}. Implemented in PyGame, it has less detailed visual effects but more complex physical dynamics, and is mainly developed to test the generalization behavior of an agent in a more challenging and controlled environment (Fig.~\ref{fig_FlappyBall}). The agent, controlled by a neural network, aims to navigate through the openings between pipes without hitting them for a certain length of time. The agent can control two actions, i.e., flapping forward and upward; both actions can be applied simultaneously. If they are not applied, gravity will pull the agent down and drag will slow it down.
The agent gets a reward of +1 every time it passes a pipe successfully, and various penalties depending how badly it crashes into the pipes, ceiling, or ground. Each time step spent in a collision incurs a penalty of -1, and in hitting the ceiling or the ground, of -5.
This way, attempting to fly through the pipes but failing is penalized less than flying through a pipe or not trying.

At every time step, the agent receives sensory information as a vector of six numerical values as indicated in Fig. \ref{fig_FlappyBall}: the vertical position of the agent (y), its horizontal and vertical velocities (v\textsubscript{x} and v\textsubscript{y}, respectively), the horizontal distance of the agent to the right edge of the closest pipe (x), and the height of the top and bottom pipes (h\textsubscript{top} and h\textsubscript{bottom}, respectively). These values are normalized to the range [0,1]. In an environment with known physical effects, this setup is a Markov Decision Process (MDP) since all the state information necessary to decide on the right action is provided to the agent. However, the effect of the agent's actions, i.e.\ flap up or forward, as well as the physical forces acting upon the agent, i.e.\ gravity and drag, can change between episodes unbeknownst to the agent, establishing a new task for the agent. Therefore, in order to perform well in new tasks, the agent has to infer such variations from its interactions with the environment over time, which makes the problem partially observable. Since acceleration and velocity are linearly correlated, such learning is possible. 

This domain is more complex than the common Flappy Bird game, which does not have the forward flap action or drag. In order to pass more pipes without a collision, the agent needs to use the forward flapping action carefully because the only way to slow down is through drag. It also needs to be cautious because it can only observe the closest pipe. By changing the effects of actions and the forces of gravity and drag, new and more challenging situations can be created, testing the generalization performance of the agent's control policy.

The Flappy Ball domain can be seen as a proxy for control and decision making problems where the changes in the environment require immediate adaptation, such as operating a vehicle under different weather conditions, configuration changes, wear and tear, or sensor malfunctions. The challenge is to adapt the existing policies to the new conditions immediately without further training, i.e. to generalize the known behavior to unseen situations.

\subsection{Evolutionary Multi-objective Optimization (EMO)}

The original Flappy Bird game is usually treated as a single-objective optimization problem, where the number of pipes passed until one is hit is maximized.  To provide for more varied behaviors, Flappy Ball is formulated as a multi-objective optimization problem instead. The number of successfully passed pipes ($f_p$) is maximized, whereas the number of any type of collisions ($f_h$, where $h$ stands for hits) is minimized. 

Non-dominated sorting genetic algorithm (NSGA-II) \citep{Deb02_NSGA2} was implemented in DEAP \citep{DEAP} as the EMO method for Flappy Ball. Although finding the safest solution ($f_h=0$) is the ultimate goal, as in the single-objective case, the diversity resulting from the multiobjective search speeds up training and helps discover well-performing solutions \citep{Knowles01_LocalOpt}. EMO algorithms use Pareto dominance to sort the solutions into sets of equally preferable solutions (or Pareto fronts). The one containing the non-dominated solutions are called Pareto-optimal set \cite{Deb_Book}; it is up to the user to select one of them based on his or her needs. In the experiments in this paper, one that is perfectly safe or close to it is usually selected.

\begin{figure*}
\begin{minipage}{0.3\textwidth}
\centering
\includegraphics[height=1.5in]{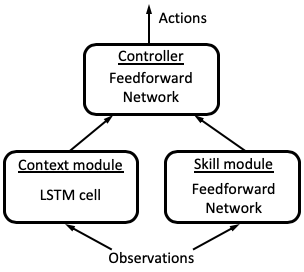}\\
(a) Context-Skill Network, CS
\end{minipage}
\hfill
\begin{minipage}{0.25\textwidth}
\centering
\includegraphics[height=1.5in]{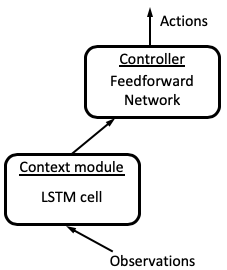}\\
(b) Context-only Network, C
\end{minipage}
\hfill
\begin{minipage}{0.3\textwidth}
\centering
\includegraphics[height=1.5in]{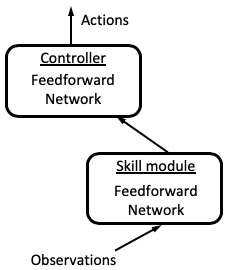}\\
(c) Skill-only Network, S
\end{minipage}
\caption{The architecture of the Context+Skill network and its ablations. (a) The network consists of three components: a Skill module that processes the current situation, a Context module that integrates observations over the entire task, and a Controller that combines the outputs of both modules, thereby using context to modulate actions. This architecture is compared to (b) context-only ablation, and (c) skill only ablation in the experiments. Each component is found to play an important role, allowing the CS network to generalize much better than its ablations.}
\label{fig_CS}
\end{figure*}

\subsection{Neural Networks} \label{sec_net}

The Context+Skill Network consists of three components: the Skill and the Context modules and the Controller (Figure~\ref{fig_CS}). The first two modules receive sensory information from the environment as numerical values, as described in Section~\ref{Game}. They send their output to the Controller, a fully connected feedforward neural network that makes the decisions on which actions to take.

The Skill module is also a fully connected feedforward network. Together with the Controller they form the Skill-only Network, S (Fig. \ref{fig_CS}(c)). The Skill module used in this study has 10 hidden and five output nodes and the Controller has 20 hidden hidden nodes and two outputs, i.e.\ one for each action. Both C and S modules have hyperbolic tangent activation function at their output layers. S is used as the baseline model throughout the study. In principle it has all the information for navigating through the pipes, but does not have the benefit of explicit representation of context.


The other main component in the Context-Skill framework is the Context module. It is composed of a vanilla Long Short Term Memory (LSTM) cell (\cite{Hochreiter97_LSTM}). There are three gates in this recurrent memory cell: input, forget, and output. The gates are responsible for learning what to store, what to throw away, and what to read out from the long-term memory of the cell. Thus, the cell can learn to retain information from the past, update it, and output it at an appropriate time, thereby making it possible to learn sequential behavior \cite{Greff17_LSTM,Geron17_MLbook}.

The C-module used in this study consists of an LSTM cell size of 10. The memory of the C-module (h\textsubscript{t-1} and c\textsubscript{t-1}) is reset at the beginning of each new task, and accumulated (transferred) across episodes within each task. It can therefore form a representation of how actions affect the environment. The output of the LSTM (h\textsubscript{t}) is sent to Controller as the context. Together the C-module and the Controller form the Context-only network, C~Fig. \ref{fig_CS}(b). It serves as a second baseline, allowing integration of observations over time, but without a specific Skill network to map them directly to action recommendations.

The complete Context-Skill Network, CS (Fig.~\ref{fig_CS}(a)) consists of both the Context and Skill modules as well as the Controller network of the same size as in C and S. The motivation behind the CS architecture, i.e.\ of integrating the Context module into S, is to make it possible for the system to learn to use an explicit context representation to modulate its actions appropriately. The method for discovering these behaviors is discussed next.

\subsection{Neuroevolution}

All three neural network models described in Section \ref{sec_net} are evolved using NSGA-II \citep{Deb02_NSGA2}. The overall procedure is shown in Algorithm~\ref{algo_evo}. The network architectures remain fixed while their weights are evolved. The goal is to maximize their average score across multiple tasks, where each task is based on different physical parameters of the Flappy Ball environment. The base values for the four actions are chosen as Flap\textsubscript{base}=-12.0 (negative value is due to the coordinate system), Gravity\textsubscript{base}=1.0, Fwd\textsubscript{base}=5.0 and Drag\textsubscript{base}=1.0. In each task during evolution, only one parameter is subject to change, while the rest are fixed at their base values. There are four tasks used in evolution, defined as:

\begin{itemize}
\item \textbf{Task-1}: The effect of the Flap action varies $\mp 20\%$ of its base value, i.e., [Flap\textsubscript{L}, Flap\textsubscript{U}] = [-14.4, -9.6];
\item \textbf{Task-2}: The effect of the Gravity force varies $\mp 20\%$ of its base value, i.e., [Gravity\textsubscript{L}, Gravity\textsubscript{U}] = [0.8, 1.2];
\item \textbf{Task-3}: The effect of the Forward action varies $\mp 20\%$ of its base value, i.e., [Forward\textsubscript{L}, Forward\textsubscript{U}] = [4.0, 6.0]; and
\item \textbf{Task-4}: The effect of the Drag force varies $\mp 20\%$ of its base value, i.e., [Drag\textsubscript{L}, Drag\textsubscript{U}] = [0.8, 1.2].
\end{itemize}

\begin{algorithm}
\caption{Preparation of task parameters}\label{prep_params}
\begin{algorithmic}[1]
\Procedure{prepareTaskParams}{n$_{\mathrm{episodes}}$, n$_{\mathrm{tasks}}$}
    \State p := [] \ \ \ \#Parameter vector
    \State S := randomInteger(2\textsuperscript{32}, n$_{\mathrm{episodes}}$)
    \State F := randomUniform(Flap$_{\mathrm L}$, Flap$_{\mathrm U}$, n$_{\mathrm {episodes}}$)
    \State G := randomUniform(Gravity$_{\mathrm L}$, Gravity$_{\mathrm U}$, n$_{\mathrm {episodes}}$)
    \State Fwd := randomUniform(Fwd$_{\mathrm L}$, Fwd$_{\mathrm U}$, n$_{\mathrm {episodes}}$)
    \State D := randomUniform(Drag$_{\mathrm L}$, Drag$_{\mathrm U}$, n$_{\mathrm{episodes}}$)
     \For{task \textbf{from} 1 \textbf{to} n$_{\mathrm {tasks}}$}
        \For{e \textbf{from} 1 \textbf{to} n$_{\mathrm{episodes}}$}
            \If{task == 0}
             \State \scalebox{0.9}{ p \ $\cup$ \ [S[e], F[e], G$_{\mathrm{base}}$, Fwd$_{\mathrm{base}}$, D$_{\mathrm{base}}$]}
        \ElsIf{task == 1}
             \State \scalebox{0.9}{ p \ $\cup$ \ [S[e], F$_{\mathrm{base}}$, G[e], Fwd$_{\mathrm{base}}$, D$_{\mathrm{base}}$]}
        \ElsIf{task == 2}
             \State \scalebox{0.9}{ p \ $\cup$ \ [S[e], F$_{\mathrm{base}}$, G$_{\mathrm{base}}$, Fwd[e], D$_{\mathrm{base}}$]}
        \ElsIf{task == 3}
             \State \scalebox{0.9}{ p \ $\cup$ \ [S[e], F$_{\mathrm{base}}$, G$_{\mathrm{base}}$, Fwd$_{\mathrm {base}}$, D[e]]}
         \EndIf
       \EndFor
    \EndFor
    \Return p
\EndProcedure
\end{algorithmic}
\end{algorithm}

Each task, and therefore each parameter, is uniformly sampled n\textsubscript{episodes}=5 times within the limits specified above. The fitness of every individual in the population is evaluated in parallel on the same task distribution for a fair comparison. Each episode length is fixed to 500 time steps. The seed number for the random number generator is included in the task parameters so that the distribution of the pipes can be repeated.

\begin{algorithm}
\caption{Fitness evaluation}\label{eval_fit}
\begin{algorithmic}[1]
\Procedure{ evalFitness}{ind, taskParams, n$_{\mathrm{episodes}}$} 
    \State pipes, hits = [], []
    \State net = genotypeToPhenotype(ind)
    \State n$_{\mathrm{tasks}}$ = taskParams
    \For{task \textbf{from} 1 \textbf{to} n$_{\mathrm{tasks}}$}
       \If{net \textbf{contains} C}
         \State net.h$_{\mathrm{prev}}$ = [0, 0, ..., 0]\textsubscript{1xC\textsubscript{output}}
         \State net.c$_{\mathrm{prev}}$ = [0, 0, ..., 0]\textsubscript{1xC\textsubscript{output}}
       \EndIf
       \For{e \textbf{from} 1 \textbf{to} $n_{\mathrm{episodes}}$}
          \State $f_0$, $f_1$ = FlappyBall(net, taskParams)
          \State pipes \ $\cup$ \ $f_0$
          \State hits \ $\cup$ \ $f_1$
       \EndFor  
     \EndFor
     \Return pipes$_{\mathrm{mean}}$, hits$_{\mathrm{mean}}$
\EndProcedure
\end{algorithmic}
\end{algorithm}

After the task parameters are prepared, fitness evaluation follows (Algorithm~\ref{eval_fit}). The parameters of a network are stored as an array in the individual candidate and converted to the corresponding neural network representation (Line~3) before the fitness evaluation (Line~10). The memory of Context-module in CS and C is reset at the begining of each task (Lines 7--8), and transferred from episode to episode otherwise. The average number of successfully passed pipes and collisions in each episode are returned as the two objective values to be maximized and minimized, respectively. There are a total of 20 episodes, since there are four tasks with five episodes in each.

\begin{algorithm}
\caption{Evolutionary loop for training neural networks on multiple tasks}\label{algo_evo}
\begin{algorithmic}[1]
\Procedure{evolve()}{}
   \State stop := False
   \State fitness := []
   \State parents := initializePopulation($\mu$)
   \State taskParams = prepareTaskParams(n$_{\mathrm{episodes}}$,n$_{\mathrm{tasks}}$)
   \For{ind \textbf{from} 1 \textbf{to} $\mu$}
        \State fitness $\cup$ evalFitness(parents, taskParams)
   \EndFor
   \For{gen \textbf{from} 1 \textbf{to} n$_{\mathrm{gen}}$}
   		\State offspring = tournamentSelection(parents, $\mu$)
       \For{i \textbf{from} 1 \textbf{to} $\lambda$}
             \If {random() $\leq$ p$_{\mathrm{crossover}}$}
                 \State SBX(offspring[i], offspring[i+1])
             \EndIf
             \State polynomialMutation(offspring[i])
             \State polynomialMutation(offspring[i+1])
       \EndFor
       \State params = prepareTaskParams(n$_{\mathrm{episodes}}$,n$_{\mathrm{tasks}}$)
       \For{ind \textbf{from} 1 \textbf{to} $\lambda$}
           \State fitness $\cup$ evalFitness(offspring, taskParams)
       \EndFor
       \For{j \textbf{from} 1 \textbf{to} $\lambda$}
            \If {fitness[j][0] $\geq$ pipes$_{\mathrm{max}}$}
                \If {fitness[j][1] $\leq$ hits$_{\mathrm{max}}$}
                     \State stop := True
                \EndIf
            \EndIf
       \EndFor
       \State \scalebox{0.9}{ parents := tournamentSelection(parents + offspring, $\mu$) }
       \If {stop == True}
            \State \Return parents
            \State \textbf{break}
       \EndIf
   \EndFor
\EndProcedure
\end{algorithmic}
\end{algorithm}

The overall procedure, i.e., NSGA-II applied to evolving agents in the Flappy Ball domain, is shown in Algorithm~\ref{algo_evo}. It receives n$_\text{tasks}$=4, n$_\text{episodes}$=5, perturb=0.2 (i.e., $\pm$20\%), Flap$_\text{base}$ = -12.0, Gravity$_\text{base}$ = 1.0, Forward$_\text{base}$ = 5.0, Drag$_\text{base}$ = 1.0, $\mu$ = 96, p$_\text{crossover}$ = 0.9, n$_\text{gen}$ = 2,500 as input. The population size ($\mu$) is chosen as a multiple of 24 since the fitness evaluations are distributed among 24 threads on a cluster (i.e., Dell PowerEdge M710, 2x Xeon X5675, 6 core @ 3.06GHz). The details about the genetic operators such as SBX (Simulated Binary Crossover), Polynomial Mutation, and Tournament Selection Based on Dominance can be found in the literature \citep{Deb02_NSGA2}. NSGA-II uses ($\mu$ + $\lambda$) elitist selection strategy with a bias on individuals
in lower fronts, where the Pareto-optimal front is the first front. If the individuals are located in the same front, the ones that are more distant from the others in objective space are selected to maintain the diverse set of trade-off solutions within the population.

\begin{figure*}
\begin{minipage}{0.3\textwidth}
\centering
\includegraphics[width=\textwidth]{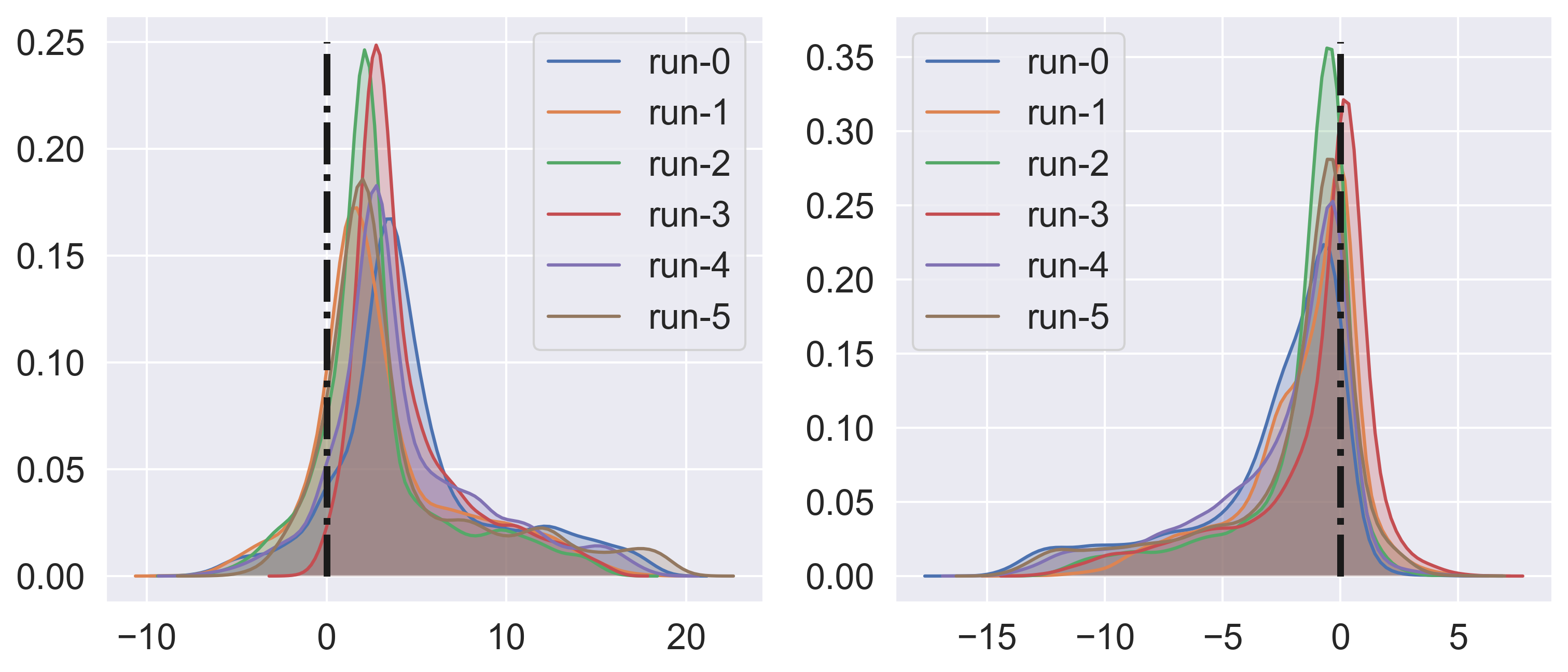}
$f_0$ (pipes)\hspace{0.25\textwidth} $f_1$ (hits)\\
(a) CS - S
\end{minipage}
\hfill
\begin{minipage}{0.3\textwidth}
\centering
\includegraphics[width=\textwidth]{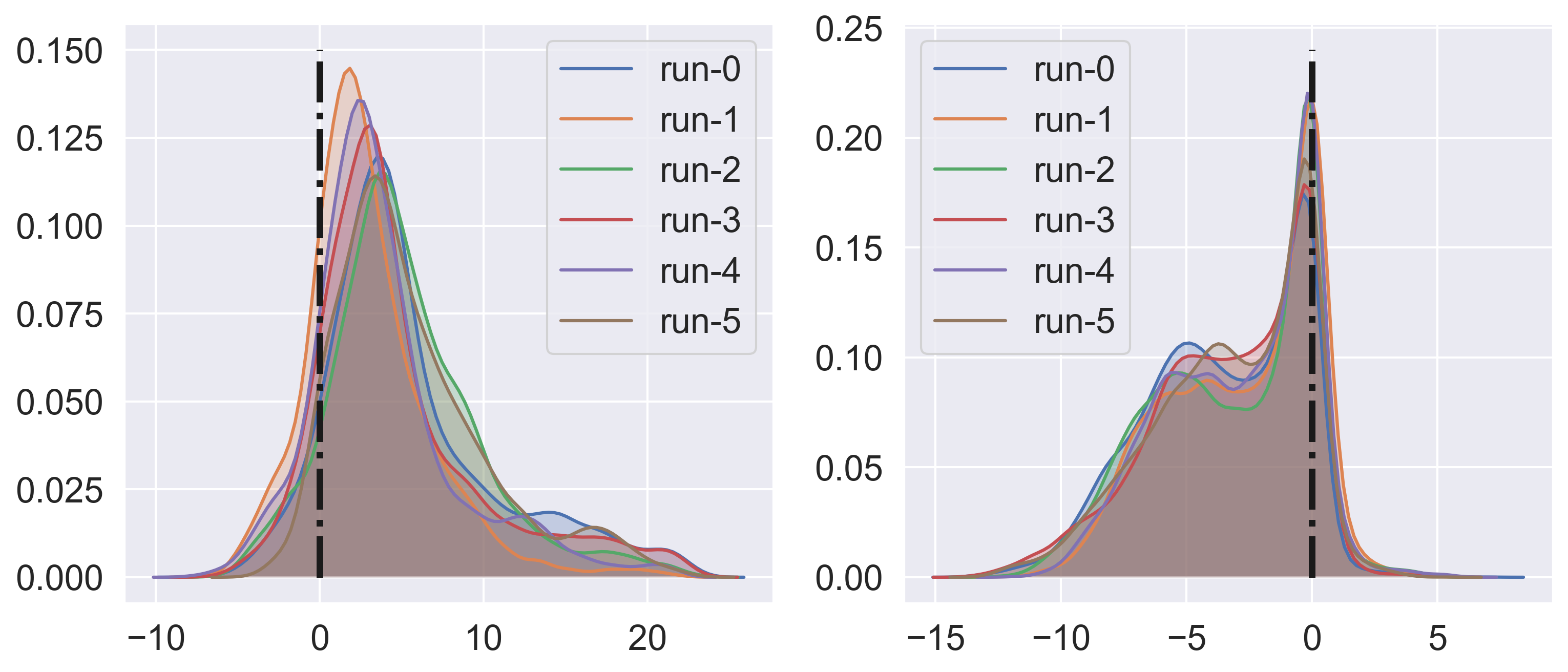}
$f_0$ (pipes)\hspace{0.25\textwidth} $f_1$ (hits)\\
(b) CS - C
\end{minipage}
\hfill
\begin{minipage}{0.3\textwidth}
\centering
\includegraphics[width=\textwidth]{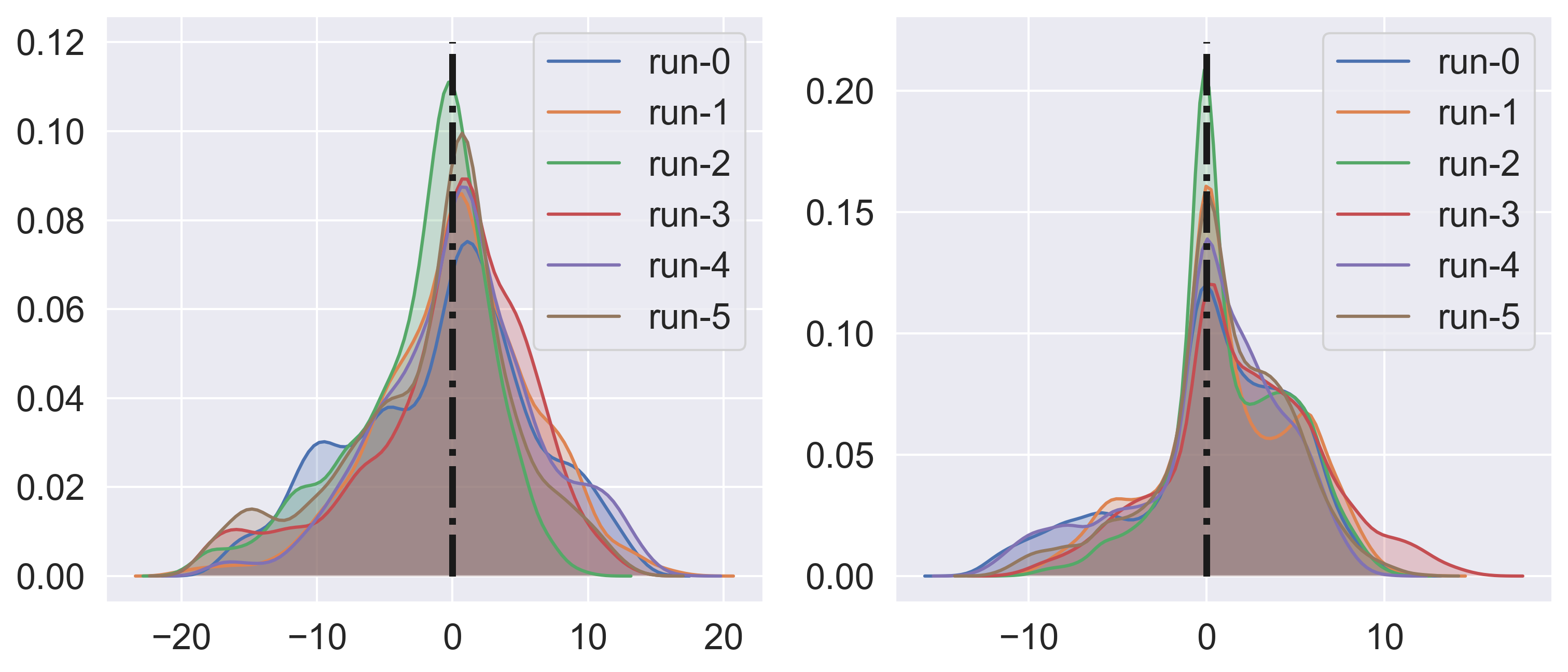}
$f_0$ (pipes)\hspace{0.25\textwidth} $f_1$ (hits)\\
(c) C - S
\end{minipage}
\caption{Generalization differences between Context+Skill network and its ablations. The $x$-axis shows the differences in generalization performance across the $3\times10^4$ test episodes for the three pairs of architectures. A distribution that is skewed to the right of the 0 line (blue dashed line) is better on the left density curve (showing $f_0$, or number of pipes), and one that is skewed to the left is better on the right curve (showing $f_1$, or number of hits). CS generalizes much better than C (a) or S (b), which are about equal (c).}
\label{histograms}
\end{figure*}

\begin{figure}
\includegraphics[width=3.25in]{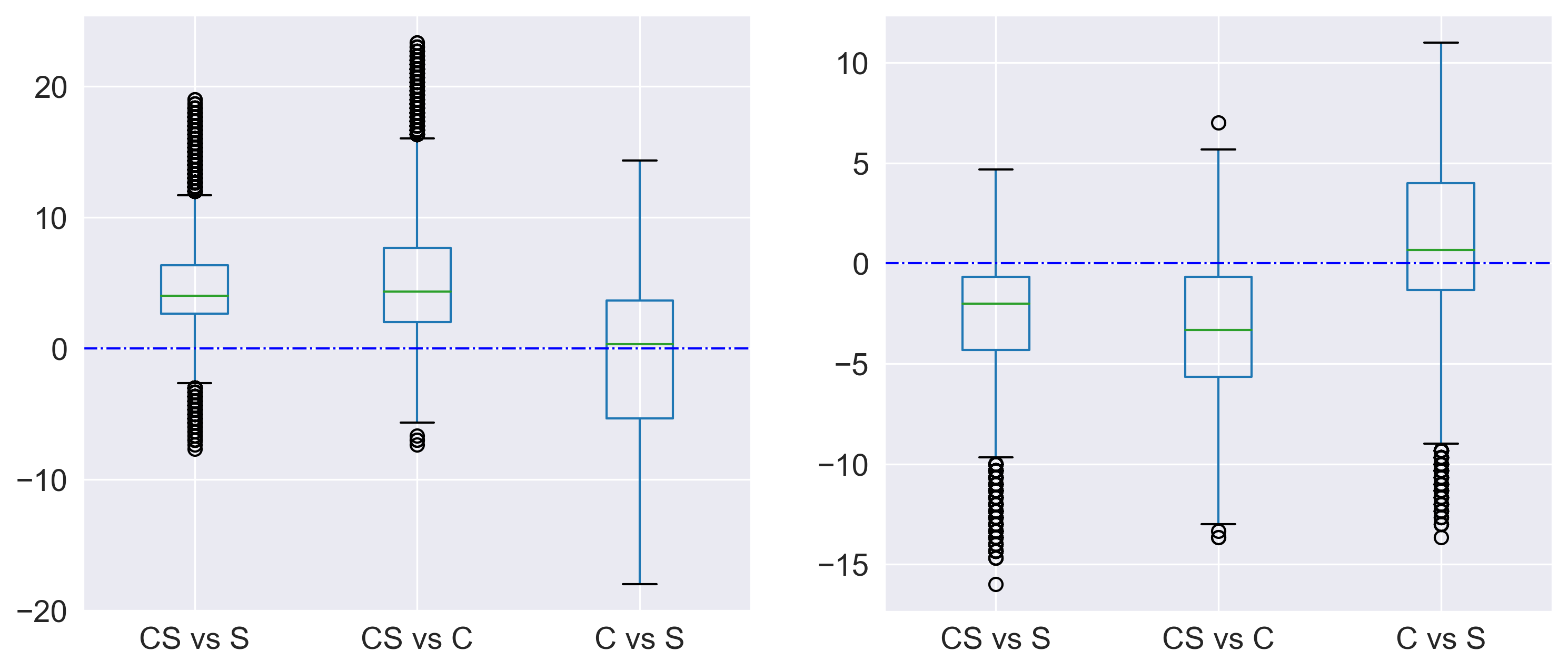}
 (a) $f_0$ (pipes) \hspace{0.15\textwidth} (b) $f_1$ (hits)\\
\caption{Summary of the generalization distributions. The data from Fig.~\ref{histograms} is organized into boxplot so that the distributions for the different architectures can be compared more clearly. CS generalizes better than both C and S, which are rather similar. CS thus combines the abilities of both C and S for superior generalization.}
\label{fig_boxplot}
\end{figure}

\section{Results} \label{sec_results}

Evolution, as given in Algorithm \ref{algo_evo}, was run 6 times with diferent random number generator seeds separately for CS, C, and S until an individual was found that achieved a fitness scores of at least $f_0$=22.0 (pipes$_{\mathrm{max}}$) and $f_1$=0.01 (hits$_{\mathrm{max}}$), where $f_0$ is the average number of successfully passed pipes and $f_1$ is the number of collisions. Although the final Pareto-optimal set in each run contained individuals with higher $f_0$ values, the minimum $f_1$ requirement meant that only relatively safe solutions were accepted.  Generalization ability of these solutions were then evaluated.

\subsection{Learning} \label{sec_learning}

The evolution of S takes the shortest amount of generations since it has the least number of model parameters to optimize, i.e., 287, compared with 982 for C and 1207 for CS.  To make sure the number of parameters was not a factor, another S with a larger Skill module, with the same number of parameters as CS, was also evolved until the same target level. However, it performed poorly compared to the smaller S in the generalization studies, apparently because it was easier to overfit. Thus, it was excluded from the comparisons that follow.

\subsection{Generalization Behavior} \label{sec_generalization}

To evaluate the generalization performance of the best performing networks, the task parameters (i.e., flap, gravity, forward, and drag) were changed in the following two ways while keeping the networks fixed:

\begin{itemize}
\item The range of variation in the task parameters was increased from 20\% to \textbf{75\%}; and
\item All four parameters were varied \textbf{simultaneously} as opposed to one at a time.
\end{itemize}

The task parameters were varied in a four-dimensional structured grid ranging from each parameter's 25\% and 175\% of the base value, respectively. Thus, with the updated limits, the effect of

\begin{itemize}
\item the flap action varied between [-21.0, -3]\textsubscript{test};
\item the gravity force varied between [0.25, 1.75]\textsubscript{test};
\item the forward action varied between [1.25, 8.75]\textsubscript{test}; and
\item the drag force varied between [0.25, 1.75]\textsubscript{test}.
\end{itemize}

Each parameter axis was divided into 10 equal steps and each set of task parameters were sampled three times (with varying pipe distribution) and averaged. Therefore, all three networks were tested for $3\times10^4$ episodes. To compare the generalization performance of the networks pairwise, the difference in the number of successfully passed pipes and the number of collisions are presented in the following density plots of Figure~\ref{histograms}, where all (6) of the density curves visibly overlap, given the stochasticity of evolutionary algorithms. The horizontal axis shows the difference in either $f_0$ or $f_1$, whereas the vertical axis shows the probability of these results. Having a skewed distribution to the right side of the 0-value is better for the left histogram (i.e., score of pipes), whereas the opposite is better for the right histogram (i.e., score of hits) for each network.

\begin{figure}
\includegraphics[width=3.3in]{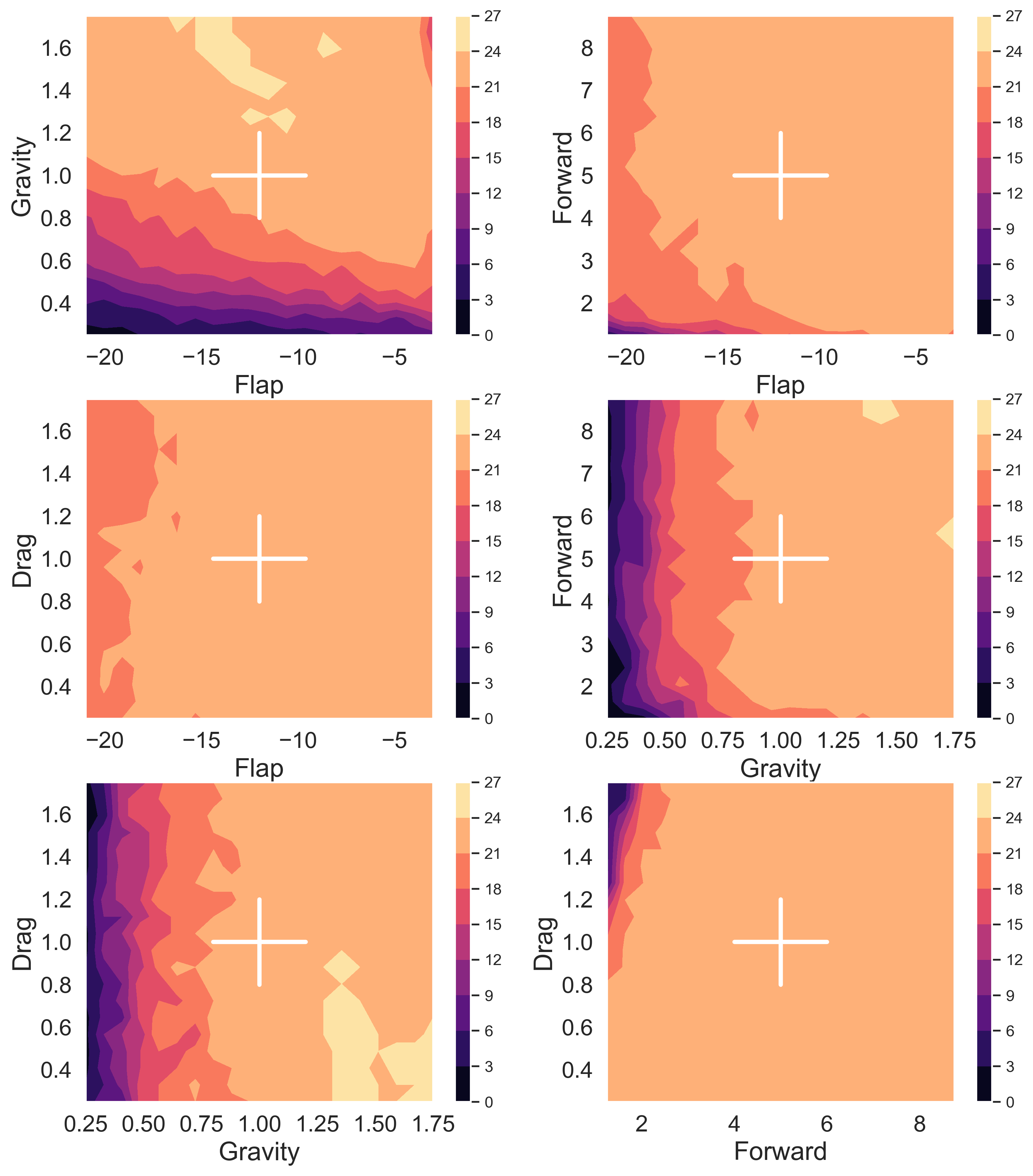}
\caption{Generalization ability of CS (i.e.\ f\textsubscript{pipes}) in 2D-parameter space, where all pairwise combinations are tested. The white cross in each contour plot indicates interpolation axes and they show variation during training. It is visible that the parameter range used in testing is much wider. It is remarkable to see that CS can achieve a maximum score of 27 pipes and an average score of 21 pipes.}
\label{fig_contourplots}
\end{figure}

The density plots show that CS performs better than both C and S by a large margin (Fig.~\ref{histograms}(a) and~(b)). Interestingly, C and S have similar generalization even though they have very different architectures (Fig.~\ref{histograms}(c)). These results are also evident in the summary boxplot of Fig.~\ref{fig_boxplot}. The contour plots in Fig.~\ref{fig_contourplots} makes the generalization capability of CS more visible in two parameter space where all pairwise parameter combinations are tested in a similar way. The white cross in each contour plot indicates interpolation axes to show variation during training. Therefore, even though each of C and S do not perform well alone, when combined into CS, they work well together and allow generalization to a wide range of new situations.


\begin{figure*}
\begin{minipage}{0.3\textwidth}
\centering
\includegraphics[height=1.85in]{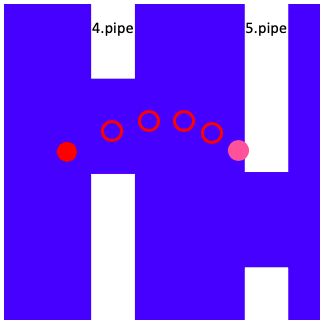}\\
(a) Skill-only Network, S
\end{minipage}
\hfill
\begin{minipage}{0.3\textwidth}
\centering
\includegraphics[height=1.85in]{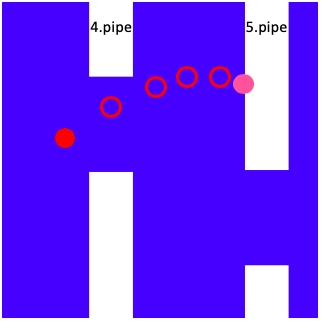}\\
(b) Context-only Network, C
\end{minipage}
\hfill
\begin{minipage}{0.3\textwidth}
\centering
\includegraphics[height=1.85in]{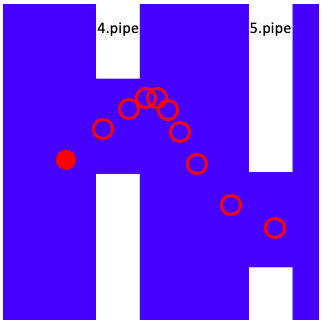}\\
(c) Context-Skill Network, CS
\end{minipage}
\caption{Contrasting the generalization ability of (a) S, (b) C, and (c) CS. At the 4th pipe, S and C flap up and then forward, end up too high too fast without enough time to come back down, and crashing into the 5th pipe. In contrast, the Context-Skill Network avoids the collision by correctly estimating the effects of its actions, giving itself enough time to come down. For an animation of these episodes, see https://drive.google.com/drive/folders/1GBdJzD9tDHJkd59YbQUOIQua6nCiLjXa.}
\label{fig_crash}
\end{figure*}

\section{Behavior Analysis}
\label{behavior}

To understand how the CS architecture outperforms its individual components C and S, a set of task parameters [Flap=-7.0, Gravity=0.58, Fwd=8.75, Drag=0.58], which was included in the generatization tests presented in the Results section was evaluated further. This setting has previously unseen exaggarated effects for flap and forward, and previously unseen diminished effects for gravity and drag. Thus, actions tend to push up and speed up the agents more than expected, and it is difficult for it to slow down and come down. Generalization requires both extrapolation of the task parameter limits as well as understanding previously unseen interaction between them. All three networks were tested in the same environment and their behavior tracked in detail.

The C network was able to pass 15 pipes successfully, and collided with six pipes, whereas S performed slightly better by passing 16 pipes with five collisions. On the other hand, CS remarkably managed to pass all 21 pipes without hitting any of them. In this particular test case, both C and S use all four actions (flap, forward, simultaneously flap and forward, or do nothing, i.e.\ glide), but CS interestingly never uses flap. That action simply lifts the agent up, which is rarely optimal action in this environment where it takes such a long time to come down. If it is necessary to go up that is because the opening is high, and in that case it is more efficient to move forward as well.

As an illustration, Fig.~\ref{fig_crash} shows a situation at the 4th and 5th pipe. Both C and S make a similar mistake by flapping up and forward. They end up too high too fast, do not have enough time to come back down, and crash into the 5th pipe. In contrast, as soon as the 5th pipe becomes visible, CS refrains from both actions while there is enough time for weaker gravity and drag to slow and pull down the agent, and it reaches the opening in the 5th pipe just fine.

\section{Discussion and Future Work}
The proposed Context+Skill approach adapts to unseen situations by representing context explicitly. Compared to its components, it has a remarkable ability to generalize to unseen situations. In this proof-of-concept study, the architecture of the neural network model has a fixed-topology which constrains the model's functionality. Evolution of the network topology together with its weights \citep{Stanley02_NEAT, Schrum14_Modular} will be a natural extension to this work. 

Besides the architecture, the choice of the tasks used for training plays an important role in the generalization capability of the model. Therefore, one direction for future work is to investigate methodologies that can automatically design a curriculum, i.e., a set of new training tasks and a better order to learn them \citep{Narvekar18_CL, Wang19_POET, Schmidhuber11_Powerplay, Justesen18_AutoCL, Risi19_PCG}.

In addition to Flappy Ball experiments, the Context+Skill model will be tested in more environments similar to Flappy Ball domain (e.g., LunarLander-v2) and the autonomous driving simulation, CARLA \citep{Dosovitskiy17_CARLA}. Instead of using handcrafted features, convolutional layers to be added in front of the Context and Skill modules can be used to discover features on its own while training. This would enrich the variety of tasks that the model can be tested on \citep{CoinRun, AnimalAI}.

Another direction for future work is to look into the hidden layer patterns to see if any evidence can be found for the observed generalization capabilities \citep{Zhang16_RethinkingGeneralization} or representational capacity \citep{Arpit17_MemorizationDNNs}. There is plenty of work about learned hierarchical representations in applications such as computer vision \citep{Yosinski15_DeepVisualization} and natural language understanding, however it is still limited in reinforcement learning tasks \citep{Karpathy_Unreasonable}.

Lifelong machine learning tries to mimic how humans and animals learn by accumulating the knowledge gained from past experience and using it to incrementally adapt to new situations \citep{Parisi18_Continual}. The generalization ability presented in this work can serve as a foundation for continual learning. It can provide an initial rapid adaptation to new situations upon which further learning can be based. How to convert generalization into a permanent ability in this manner is an interesting direction of future research.

\section{Conclusion}

Perhaps the main challenge in deploying artificial agents in the real world is that they are brittle---they can only perform well in situations for which they were trained. However, this paper demonstrates an alternative approach based on separating contexts from the actual skills. Context can then be used to modulate the actions in a systematic manner, significantly extending the unseen situations that can be handled. This principle was evaluated in a challenging version of the Flappy Bird game, and shows to perform better than traditional training and general memory-based training. This Context+Skill approach should be useful in many control and decision making tasks in the real world. 

\section{Acknowledgments}

This research was supported in part by DARPA L2M Award DBI-0939454.


\footnotesize
\bibliographystyle{apalike}
\bibliography{alife_ref} 

\begin{thebibliography}{}

\bibitem[Arpit et~al., 2017]{Arpit17_MemorizationDNNs}
Arpit, D., Jastrzębski, S., Ballas, N., Krueger, D., Bengio, E., Kanwal,
  M.~S., Maharaj, T., Fischer, A., Courville, A., Bengio, Y., and
  Lacoste-Julien, S. (2017).
\newblock A closer look at memorization in deep networks.

\bibitem[Beyret et~al., 2019]{AnimalAI}
Beyret, B., Hern{\'{a}}ndez{-}Orallo, J., Cheke, L., Halina, M., Shanahan, M.,
  and Crosby, M. (2019).
\newblock The animal-ai environment: Training and testing animal-like
  artificial cognition.
\newblock {\em CoRR}, abs/1909.07483.

\bibitem[Cobbe et~al., 2019]{CoinRun}
Cobbe, K., Klimov, O., Hesse, C., Kim, T., and Schulman, J. (2019).
\newblock Quantifying generalization in reinforcement learning.
\newblock In {\em Proceedings of the 36th International Conference on Machine
  Learning, {ICML} 2019, 9-15 June 2019, Long Beach, California, {USA}},
  volume~97 of {\em Proceedings of Machine Learning Research}, pages
  1282--1289.

\bibitem[Deb, 2001]{Deb_Book}
Deb, K. (2001).
\newblock {\em Multi-Objective Optimization Using Evolutionary Algorithms}.
\newblock John Wiley \& Sons, Inc., USA.

\bibitem[Deb et~al., 2002]{Deb02_NSGA2}
Deb, K., Pratap, A., Agarwal, S., and Meyarivan, T. (2002).
\newblock A fast and elitist multiobjective genetic algorithm: Nsga-ii.
\newblock {\em IEEE Transactions on Evolutionary Computation}, 6(2):182--197.

\bibitem[Dosovitskiy et~al., 2017]{Dosovitskiy17_CARLA}
Dosovitskiy, A., Ros, G., Codevilla, F., Lopez, A., and Koltun, V. (2017).
\newblock {CARLA}: {An} open urban driving simulator.
\newblock In {\em Proceedings of the 1st Annual Conference on Robot Learning},
  pages 1--16.

\bibitem[Fernando et~al., 2018]{Fernando18_Baldwin}
Fernando, C., Sygnowski, J., Osindero, S., Wang, J., Schaul, T., Teplyashin,
  D., Sprechmann, P., Pritzel, A., and Rusu, A. (2018).
\newblock Meta-learning by the baldwin effect.
\newblock {\em Proceedings of the Genetic and Evolutionary Computation
  Conference Companion}.

\bibitem[Finn et~al., 2017]{Finn17_MAML}
Finn, C., Abbeel, P., and Levine, S. (2017).
\newblock Model-agnostic meta-learning for fast adaptation of deep networks.
\newblock In {\em Proceedings of the 34th International Conference on Machine
  Learning - Volume 70}, ICML’17, page 1126–1135. JMLR.org.

\bibitem[Fortin et~al., 2012]{DEAP}
Fortin, F.-A., {De Rainville}, F.-M., Gardner, M.-A., Parizeau, M., and
  Gagn\'e, C. (2012).
\newblock {DEAP}: Evolutionary algorithms made easy.
\newblock {\em Journal of Machine Learning Research}, 13:2171--2175.

\bibitem[G\'eron, 2017]{Geron17_MLbook}
G\'eron, A. (2017).
\newblock {\em Hands-On Machine Learning with Scikit-Learn and TensorFlow:
  Concepts, Tools, and Techniques to Build Intelligent Systems}.
\newblock O’Reilly Media, Inc., 1st edition.

\bibitem[Grbic and Risi, 2019]{Grbic19_EvoMetaLearn}
Grbic, D. and Risi, S. (2019).
\newblock Towards continual reinforcement learning through evolutionary
  meta-learning.
\newblock In {\em Proceedings of the Genetic and Evolutionary Computation
  Conference Companion}, GECCO '19, pages 119--120. Association for Computing
  Machinery.

\bibitem[Greff et~al., 2017]{Greff17_LSTM}
Greff, K., Srivastava, R.~K., Koutnik, J., Steunebrink, B.~R., and Schmidhuber,
  J. (2017).
\newblock Lstm: A search space odyssey.
\newblock {\em IEEE Transactions on Neural Networks and Learning Systems},
  28(10):2222--2232.

\bibitem[Hochreiter and Schmidhuber, 1997]{Hochreiter97_LSTM}
Hochreiter, S. and Schmidhuber, J. (1997).
\newblock Long short-term memory.
\newblock {\em Neural Comput.}, 9(8):1735--1780.

\bibitem[{Justesen} and {Risi}, 2018]{Justesen18_AutoCL}
{Justesen}, N. and {Risi}, S. (2018).
\newblock Automated curriculum learning by rewarding temporally rare events.
\newblock In {\em 2018 IEEE Conference on Computational Intelligence and Games
  (CIG)}, pages 1--8.

\bibitem[Kansky et~al., 2017]{Kansky17_Schema}
Kansky, K., Silver, T., M{\'e}ly, D.~A., Eldawy, M., L{\'a}zaro-Gredilla, M.,
  Lou, X., Dorfman, N., Sidor, S., Phoenix, S., and George, D. (2017).
\newblock Schema {n}etworks: Zero-shot transfer with a generative causal model
  of intuitive physics.
\newblock In {\em Proceedings of the 34th International Conference on Machine
  Learning-Volume 70}, pages 1809--1818. JMLR. org.

\bibitem[Karpathy, 2015]{Karpathy_Unreasonable}
Karpathy, A. (2015).
\newblock The unreasonable effectiveness of recurrent neural networks.
\newblock http://karpathy.github.io/2015/05/21/rnn-effectiveness/.

\bibitem[Knowles et~al., 2001]{Knowles01_LocalOpt}
Knowles, J.~D., Watson, R.~A., and Corne, D.~W. (2001).
\newblock Reducing local optima in single-objective problems by
  multi-objectivization.
\newblock In Zitzler, E., Thiele, L., Deb, K., Coello~Coello, C.~A., and Corne,
  D., editors, {\em Evolutionary Multi-Criterion Optimization}, pages 269--283.
  Springer Berlin Heidelberg.

\bibitem[Li and Miikkulainen, 2017]{Li17_Poker}
Li, X. and Miikkulainen, R. (2017).
\newblock Evolving adaptive poker players for effective opponent exploitation.
\newblock In {\em AAAI-17 Workshop on Computer Poker and Imperfect Information
  Games}.

\bibitem[Li and Miikkulainen, 2018]{li:gecco18}
Li, X. and Miikkulainen, R. (2018).
\newblock Opponent modeling and exploitation in poker using evolved recurrent
  neural networks.
\newblock In {\em Proceedings of The Genetic and Evolutionary Computation
  Conference (GECCO 2018)}, Kyoto, Japan. ACM.

\bibitem[Narvekar and Stone, 2019]{Narvekar18_CL}
Narvekar, S. and Stone, P. (2019).
\newblock Learning curriculum policies for reinforcement learning.
\newblock In {\em Proceedings of the 18th International Conference on
  Autonomous Agents and MultiAgent Systems}, AAMAS ’19, page 25–33,
  Richland, SC. International Foundation for Autonomous Agents and Multiagent
  Systems.

\bibitem[Parisi et~al., 2019]{Parisi18_Continual}
Parisi, G.~I., Kemker, R., Part, J.~L., Kanan, C., and Wermter, S. (2019).
\newblock Continual lifelong learning with neural networks: A review.
\newblock {\em Neural Networks}, 113:54 -- 71.

\bibitem[Risi and Togelius, 2019]{Risi19_PCG}
Risi, S. and Togelius, J. (2019).
\newblock Procedural content generation: From automatically generating game
  levels to increasing generality in machine learning.

\bibitem[Schmidhuber, 1987]{Schmidhuber87_PhD}
Schmidhuber, J. (1987).
\newblock {\em Evolutionary principles in self-referential learning, or on
  learning how to learn: The meta-meta-... hook}.
\newblock PhD thesis, Institut für Informatik, Technische Universität
  München.

\bibitem[Schmidhuber, 2011]{Schmidhuber11_Powerplay}
Schmidhuber, J. (2011).
\newblock Powerplay: Training an increasingly general problem solver by
  continually searching for the simplest still unsolvable problem.

\bibitem[Schrum and Miikkulainen, 2014]{Schrum14_Modular}
Schrum, J. and Miikkulainen, R. (2014).
\newblock Evolving multimodal behavior with modular neural networks in {M}s.
  {P}ac-{M}an.
\newblock In {\em Proceedings of the Genetic and Evolutionary Computation
  Conference Companion}, GECCO '14, pages 325--332. Association for Computing
  Machinery.

\bibitem[Stanley and Miikkulainen, 2002]{Stanley02_NEAT}
Stanley, K.~O. and Miikkulainen, R. (2002).
\newblock Evolving neural networks through augmenting topologies.
\newblock {\em Evol. Comput.}, 10(2):99--127.

\bibitem[Thrun and Pratt, 1998]{Thrun98_Learn}
Thrun, S. and Pratt, L. (1998).
\newblock {\em Learning to Learn: Introduction and Overview}, pages 3--17.
\newblock Kluwer Academic Publishers, USA.

\bibitem[Wang et~al., 2019]{Wang19_POET}
Wang, R., Lehman, J., Clune, J., and Stanley, K.~O. (2019).
\newblock Paired open-ended trailblazer ({POET}): Endlessly generating
  increasingly complex and diverse learning environments and their solutions.

\bibitem[Wikipedia, 2020]{FlappyBird_wiki}
Wikipedia ((Online; accessed 3-February-2020)).
\newblock Flappy {B}ird.
\newblock https://en.wikipedia.org/wiki/Flappy\_Bird.

\bibitem[Yosinski et~al., 2015]{Yosinski15_DeepVisualization}
Yosinski, J., Clune, J., Nguyen, A.~M., Fuchs, T.~J., and Lipson, H. (2015).
\newblock Understanding neural networks through deep visualization.
\newblock {\em CoRR}, abs/1506.06579.

\bibitem[Zhang et~al., 2016]{Zhang16_RethinkingGeneralization}
Zhang, C., Bengio, S., Hardt, M., Recht, B., and Vinyals, O. (2016).
\newblock Understanding deep learning requires rethinking generalization.
\newblock cite arxiv:1611.03530Comment: Published in ICLR 2017.

\end{thebibliography}

\end{document}